\documentclass{article}

\usepackage[english]{babel}

\usepackage[letterpaper,top=2cm,bottom=2cm,left=3cm,right=3cm,marginparwidth=1.75cm]{geometry}

\usepackage{amsmath}
\usepackage{graphicx}
\usepackage{enumitem}
\usepackage[colorlinks=true, allcolors=blue]{hyperref}
\usepackage{listings}
\usepackage{overpic}
\usepackage{ragged2e}
\usepackage{caption}
\usepackage{subcaption}
\usepackage[style=ieee, sorting=none, maxbibnames=99]{biblatex}
\usepackage{verbatim}
\usepackage{gensymb}
\usepackage{algorithm}
\usepackage[noend]{algpseudocode}

\usepackage{amssymb}
\usepackage{authblk}
\usepackage{listings}
\usepackage{color}
\usepackage{xcolor}

\definecolor{codebg}{rgb}{0.97,0.97,0.97}
\definecolor{keywordcolor}{rgb}{0.5,0.0,0.5}    
\definecolor{operatorcolor}{rgb}{0.8,0.0,0.8}   

\usepackage{textcomp,xcolor}
\definecolor{MatlabCellColour}{RGB}{250,250,250}
\definecolor{MatPurp}{rgb}{.625,.1406,.9375}
\usepackage{listings}

 \lstdefinestyle{customP}{
  belowcaptionskip=.25\baselineskip,
  breaklines=true,
  frame=L,
  xleftmargin=\parindent,
  language=Python,
  showstringspaces=false,
  basicstyle=\small\ttfamily,
  keywordstyle=\bfseries\color{white!30!black},
  identifierstyle=\color{blue},  
  commentstyle=\itshape\color{green!60!black},
  stringstyle=\color{MatPurp},
  backgroundcolor=\color{MatlabCellColour}
 }
\makeatletter
\newcommand{\longdash}[1][2em]{%
  \makebox[#1]{$\m@th\smash-\mkern-7mu\cleaders\hbox{$\mkern-2mu\smash-\mkern-2mu$}\hfill\mkern-7mu\smash-$}}
\makeatother
\newcommand{\omitskip}{\kern-\arraycolsep}

\DeclareMathOperator*{\argmax}{argmax}

\newcommand{\ba}{\mathbf{a}}
\newcommand{\bA}{\mathbf{A}}

\newcommand{\Sel}{\mathbb{S}}
\newcommand{\bTi}{\mathbf{\Psi}} 

\newcommand{\bx}{\mathbf{x}}
\newcommand{\by}{\mathbf{y}}

\newcommand{\be}{\mathbf{e}}
\newcommand{\reals}{\mathbb{R}}

\DeclareRobustCommand{\okina}{%
	\raisebox{\dimexpr\fontcharht\font`A-\height}{%
		\scalebox{0.8}{`}~%
	}%
}

\title{PySensors 2.0: A Python Package for Sparse Sensor
Placement}
\date{}
\author[1]{Niharika Karnik\footnote{nkarnik@uw.edu}}
\author[1]{Yash Bhangale\footnote{yash6599@uw.edu}}
\author[2]{Mohammad G. Abdo}
\author[3]{Andrei A. Klishin}
\author[2]{Joshua J. Cogliati}
\author[6]{Bingni W. Brunton}
\author[7]{J. Nathan Kutz}
\author[1]{Steven L. Brunton}
\author[1]{Krithika Manohar\footnote{kmanohar@uw.edu}}

\affil[1]{Department of Mechanical Engineering, University of Washington}
\affil[2]{Idaho National Laboratory}
\affil[3]{Department of Mechanical Engineering, University of Hawai\okina i at Mānoa}
\affil[6]{Department of Biology, University of Washington}
\affil[7]{Department of Applied Mathematics, University of Washington}
\begin{document}
\maketitle
\begin{abstract}
    \texttt{PySensors} is a Python package for selecting and placing a sparse set of sensors for reconstruction and classification tasks. In this major update to \texttt{PySensors}, we introduce spatially constrained sensor placement capabilities, allowing users to enforce constraints such as maximum or exact sensor counts in specific regions, incorporate predetermined sensor locations, and maintain minimum distances between sensors. We extend functionality to support custom basis inputs, enabling integration of any data-driven or spectral basis. We also propose a thermodynamic approach that goes beyond a single ``optimal'' sensor configuration and maps the complete landscape of sensor interactions induced by the training data. This comprehensive view facilitates integration with external selection criteria and enables assessment of sensor replacement impacts. The new optimization technique also accounts for over- and under-sampling of sensors, utilizing a regularized least squares approach for robust reconstruction. Additionally, we incorporate noise-induced uncertainty quantification of the estimation error and provide visual uncertainty heat maps to guide deployment decisions. To highlight these additions, we provide a brief description of the mathematical algorithms and theory underlying these new capabilities. We demonstrate the usage of new features with illustrative code examples and include practical advice for implementation across various application domains. Finally, we outline a roadmap of potential extensions to further enhance the package's functionality and applicability to emerging sensing challenges.
\end{abstract}
\section{Introduction}
Sensor placement is critical for efficient monitoring, control, and decision-making in modern engineering systems. Sensors play a crucial role in characterizing spatio-temporal dynamics in high-dimensional, non-linear systems such as fluid flows~\cite{erichson2020shallow}, manufacturing \cite{manohar2018predicting}, geophysical~\cite{alonso2010novel} and nuclear systems \cite{karnik2024constrained}. Optimal sensor placement ensures accurate, real-time tracking of key system variables with minimal hardware and enables cost-effective, real-time system analysis and control. In general, sensor placement optimization is NP-hard and cannot be solved in polynomial time.  There are ${n \choose p} = n!/((n-p)!p!)$ possible combinations of choosing $p$ sensors from an $n$-dimensional state.
Common approaches to optimizing sensor placement include maximizing the information criteria~\cite{krause2008near}, Bayesian Optimal Experimental Design~\cite{alexanderian2021optimal}, compressed sensing~\cite{donoho2006compressed}, and heuristic methods. Many sensor placement methods have submodular objective form, which sets guarantees on how close an efficient greedy placement can be to the unknown true optimum~\cite{summers2015submodularity}.
%
Sub-modular objectives can be efficiently optimized for hundreds or thousands of candidate locations using convex~\cite{joshi2008sensor} or greedy optimization approaches~\cite{summers2015submodularity} .

\texttt{PySensors} is a Python package~\cite{de2021pysensors} dedicated to solving the complex challenge of optimal sensor placement in data-driven systems. It implements advanced sparse optimization algorithms that use dimensionality reduction techniques to identify the most informative measurement locations with remarkable efficiency~\cite{manohar2018data,brunton2016sparse,clark2020multi}. It helps users identify the best locations for sensors when working with complex high dimensional data, focusing on both reconstruction~\cite{manohar2018data} and classification~\cite{brunton2016sparse} tasks. The package follows \texttt{scikit-learn} conventions for user-friendly access while offering advanced customization options for experienced users. Designed with researchers and practitioners in mind, \texttt{PySensors} provides open-source, accessible tools that support model discovery across various scientific applications. 

This new version of \texttt{Pysensors} focuses specifically on practical engineering applications where measurement data is inherently noisy and spatial deployment constraints are unavoidable. Key improvements include constraint-aware optimization methods that handle spatial restrictions and sensor density limitations. In addition, the framework introduces uncertainty quantification metrics that track how measurement noise propagates through reconstruction algorithms, enabling robust error estimation in sensor outputs. This version of \texttt{Pysensors} implements methodologies introduced by Klishin et al.~\cite{klishin2023data} and Karnik et al.~\cite{karnik2024constrained} to make them accessible to scientists and engineers in all domains. These enhancements transform \texttt{PySensors} from a purely academic tool into a practical platform for solving real-world sensing challenges while maintaining mathematical rigor.

\begin{figure}[t!]
    \centering
    \includegraphics[width=\linewidth]{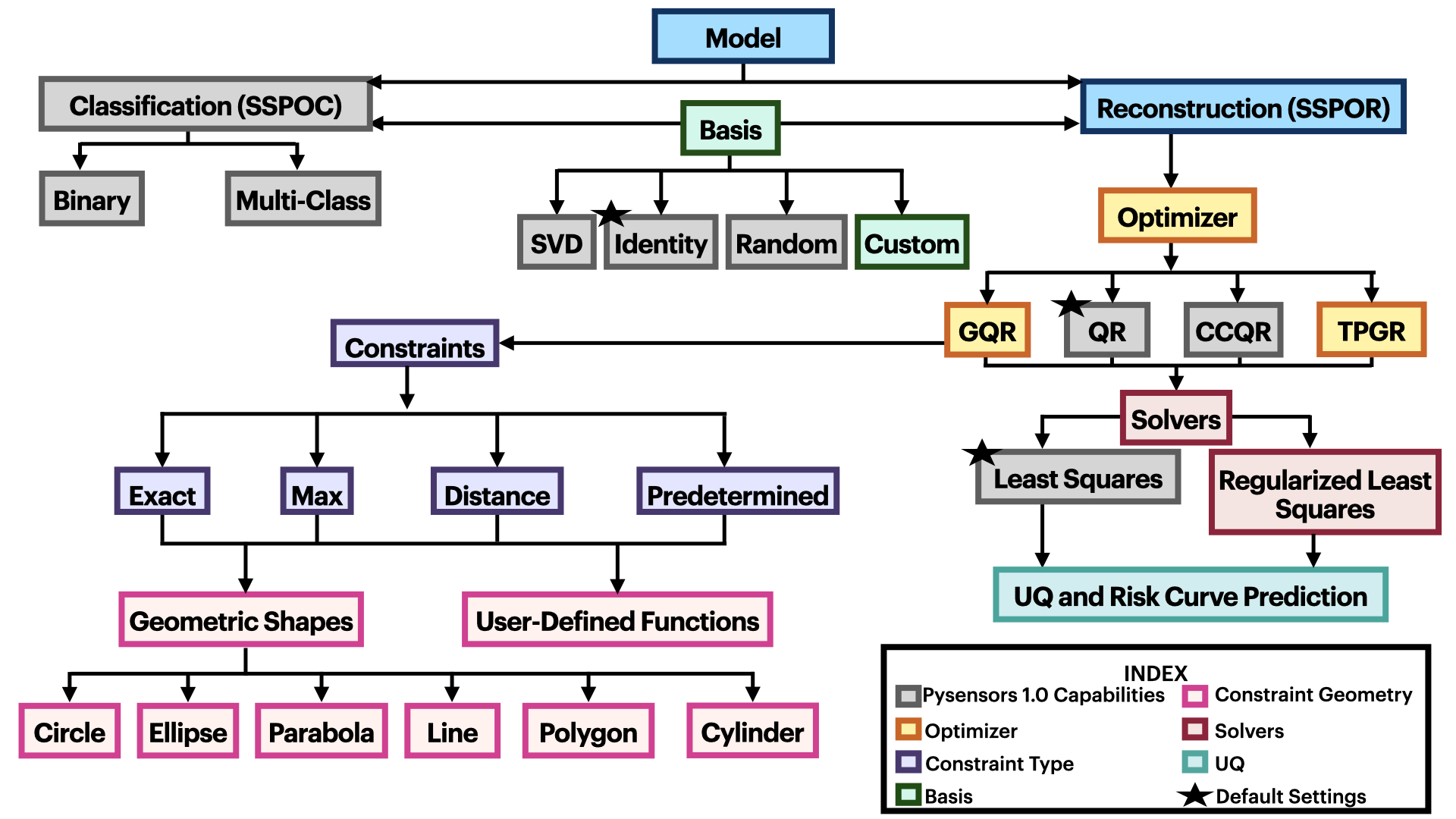}
    \caption{\texttt{PySensors 2.0} expands its capabilities by introducing custom basis functions, optimizers, constraints, solvers, and uncertainty quantification, enabling constrained sensing, over- and under-sampling,  and uncertainty quantification in the presence of noisy sensor measurements.}
    \label{fig:PysensorsOverview}
\end{figure}

 Other sensor placment packages such as \texttt{Chama}~\cite{klise2017sensor}, \texttt{Polire}~\cite{narayanan2020toolkit}, and \texttt{OSPS} toolbox~\cite{yi2011optimal}, focus primarily on event detection, Gaussian process modeling, and structural health monitoring respectively, while \texttt{PySensors} specifically targets signal reconstruction and classification applications. \texttt{PySensors 2.0} represents an advancement in sensor optimization software, establishing a distinct position in the field through its novel focus on constrained optimization and uncertainty quantification as shown in \autoref{fig:PysensorsOverview}.

In this work, we first establish the theoretical foundation of sensor placement optimization by examining mathematical techniques for signal reconstruction and classification, while introducing the dimensionality reduction approaches and optimization algorithms that underpin effective sensor selection. We then explore the novel functionalities implemented in \texttt{PySensors 2.0}, including two newly developed optimizers for hard spatial constraints for realistic deployment scenarios, adaptive strategies for sensor oversampling and undersampling, and support for custom basis functions. To demonstrate practical applications, we present a detailed case study applying our constraint-aware optimization framework to a nuclear fuel rod prototype, complete with uncertainty quantification heatmaps and estimation error metrics that provide crucial reliability insights. The paper concludes with practical implementation guidelines to help users maximize the effectiveness of these advanced sensor placement tools across diverse application domains.

\section{Background}
This section first describes the two main objectives of \texttt{PySensors}: reconstruction and classification, and their implementation using basis functions and optimization techniques for sensor placement.
\subsection{Reconstruction}
Pysensors implements \texttt{Sparse Sensor Placement Optimization for Reconstruction (SSPOR)}~\cite{manohar2018data} of full fields $\bx\in\reals^n$ from $p$ noise-corrupted sensor measurements $\by\in\reals^p$ 
\begin{equation}
    \by =  \Sel \bx + \boldsymbol\eta, 
\end{equation}
where $\boldsymbol{\eta}$ consists of zero-mean, Gaussian independent and identically distributed (i.i.d.) components, and $\Sel \in \reals^{p\times n}$ is the desired sensor (measurement) selection operator.
This measurement selection operator $\Sel$ encodes point measurements with unit entries in a sparse matrix
\begin{equation}
    \Sel = \begin{bmatrix} \be_{\gamma_1} & \be_{\gamma_2} & \hdots & \be_{\gamma_ p}\end{bmatrix}^T,
\end{equation}
where $\be_j$ are canonical basis vectors for $\mathbb{R}^n$, with a unit entry in component $j$ (where a sensor should be placed) and zeroes elsewhere. Here, $\gamma = \{\gamma_1, \gamma_2, \hdots, \gamma_p\} \subset \{1, 2, \hdots, n\}$ denotes the index set of sensor locations with cardinality $p$. 
Sensor selection then corresponds to the components of $\mathbf{x}$ that were chosen to be measured:
\begin{equation}
   \mathbf{\Sel x} = \begin{bmatrix} x_{\gamma_1} & x_{\gamma_2} & \hdots & x_{\gamma_p}\end{bmatrix}^T.
\end{equation}

The \texttt{SSPOR} class selects these sensors through a cohesive computational framework for strategically minimizing sensor deployment while maintaining reconstruction fidelity in high-dimensional systems. When initialized with measurement data $\bx\in\reals^n$, \texttt{SSPOR} first applies dimensional reduction through basis identification, then employs the computationally efficient QR factorization algorithm to determine optimal sensor locations. This approach, which has demonstrated efficacy in both reduced-order modeling and sparse sensing applications, strategically leverages the inherent structure of the identified basis to prioritize the most informative measurement points.

\subsection{Classification}
The \texttt{Sparse Sensor Placement Optimization for Classification (SSPOC)} framework identifies minimal sensor configurations that can classify high dimensional signals $\bx\in\reals^n$ as one of $c$ classes. Unlike traditional compressed sensing~\cite{baraniuk2010model,donoho2006compressed, candes2006stable} approaches that focus on signal reconstruction, \texttt{SSPOC} specifically targets the preservation of classification boundaries in feature space $\bTi_r$ by identifying the sparsest sensor set capable of reconstructing discriminant hyperplanes.

The \texttt{SSPOC} architecture accepts any combination of basis representation and linear classifier, defaulting to Linear Discriminant Analysis (LDA) and Identity basis when not otherwise specified. The optimization pipeline proceeds through multiple stages: dimensional reduction via basis fitting, classifier training within the reduced space, sparse optimization incorporating both classifier weights and basis structures, sensor selection based on optimization outputs, and optional classifier retraining using only measurements from the selected locations.

The framework adapts its optimization strategy according to classification complexity. For binary problems, it employs Orthogonal Matching Pursuit from the \texttt{scikit-learn} library, while multi-class scenarios trigger multi-task Lasso regularization. This mathematical flexibility enables the framework to address diverse classification challenges while maintaining computational efficiency and sparse sensing requirements.

\subsection{Basis}
High dimensional field dynamics $\bx\in\reals^n$ can be represented as a linear combination of spatial basis modes $\boldsymbol{\Psi}$ weighted by time-varying coefficients $\mathbf{a}$ 
$$\bx = \bTi_r\ba.$$
This basis, which can be built from spectral or data-driven decomposition methods, is typically chosen so that the embedding dimension is as small as possible, i.e., $r\ll n$. Different basis functions can significantly impact sensor selection effectiveness and reconstruction quality~\cite{manohar2018data}. PySensors offers several interchangeable \texttt{basis} options for sparse sensor selection:
\begin{enumerate}
    \item \texttt{Identity}: Processes raw measurement data directly without transformation.
    \item \texttt{SVD}: Utilizes truncated singular value decomposition's $\mathbf{X} = \mathbf{U}_r \mathbf{\Sigma}_r \mathbf{V}^*_r$ left singular vectors $\mathbf{U}_r$, computing only the specified number of modes to reduce computational demands. A randomized SVD option further enhances efficiency.
    \item \texttt{RandomProjection} : Projects measurements into a new space by multiplying them with random Gaussian vectors, connecting to compressed sensing methodologies from established literature~\cite{baraniuk2010model,donoho2006compressed, candes2006stable}.
    \item \texttt{Custom}: This is a new option included in \texttt{Pysensors 2.0} that enables users to leverage custom basis functions beyond \texttt{PySensors}' built-in options. Researchers can transform their data into an alternative basis such as dynamic mode decomposition modes \cite{schmid2010dynamic}, before passing it to a PySensors instance configured with the Custom basis.
\end{enumerate}

\subsection{Optimizers}
When a user specifies $p$ sensors, SSPOR first fits a basis $\bTi_r$ to the data and optimizes sensor placement by minimizing the following objective function:
\begin{equation}
\label{eqn:detobj}
    \gamma_* = \argmax_{\gamma, |\gamma|= p} \; \log \det((\Sel \bTi_r)^T(\Sel \bTi_r)). 
\end{equation}
where $\gamma_*$ denotes the index set of optimized sensor locations with cardinality $p$. When $p = r$,~\autoref{eqn:detobj} is equivalent to the maximizer of $\log |\det(\Sel \bTi_r)|$.
Direct optimization of this criterion leads to a brute force combinatorial search. This sensor placement approach builds upon the empirical interpolation method (EIM) \cite{barrault2004empirical} and discrete empirical interpolation method (DEIM) \cite{chaturantabut2010nonlinear} to develop a greedy strategy for optimizing sensor selection built upon the pivoted QR factorization~\cite{drmac2016new,manohar2018data, manohar2018predicting, manohar2021optimal}. 

PySensors implements the \texttt{QR} optimizer for optimal sensor selection in unconstrained scenarios where the number of sensors $p$ equals the number of modes $r$. The framework incorporates heterogeneous cost functions into the optimization process to accommodate practical deployment constraints. For example, when monitoring oceanographic parameters, the system can account for the substantially higher costs of deep-sea sensors relative to coastal installations~\cite{clark2020multi}. This capability is implemented through the \texttt{Cost-Constrained QR (CCQR)} algorithm in the optimizers submodule, allowing users to balance information capture against resource limitations when designing sensor networks for complex physical systems.

Traditional QR factorization presents challenges in under-sampling $p < r$ and over-sampling $p > r$ scenarios, as well as when spatial constraints must be considered. \texttt{PySensors 2.0} addresses these limitations through two new optimization algorithms: \texttt{Generalized QR (GQR)} and \texttt{Two Point GReedy (TPGR)}.

\section{New Functionality}
\texttt{PySensors 2.0} has been enhanced to address critical challenges in sensor placement by incorporating spatial constraints, noise-induced uncertainties, over/under sampling and sensor interactions. These advancements optimize reconstruction performance for complex processes across nuclear energy, fluid dynamics, biological systems, and manufacturing applications, enabling more accurate modeling, prediction, and control capabilities.
\subsection{Hard Constraints}
\label{sec:HardConstraints}
The previous version of \texttt{Pysensors} implements two sensor placement approaches: (1) an unconstrained optimization formulated as \texttt{QR} that allows sensors to be placed anywhere in the domain, and (2) a cost-constrained optimization formulated as \texttt{CCQR} that incorporates variable placement costs, making certain regions more expensive for sensor deployment. 

Many engineering applications have extreme operating conditions, high costs, limited accessibility and safety regulations that impose significant constraints on spatial locations of sensors. Implementing spatially constrained sensor placement requires a deeper intervention in the underlying QR optimization framework. To address this requirement, we have developed a new optimization functionality called \texttt{General QR (GQR)} based on Karnik et. al.~\cite{karnik2024constrained}, which provides the architectural flexibility needed to handle complex spatial constraints. In \texttt{Pysensors 2.0} we enhance the algorithm's capabilities by incorporating diverse spatial constraints defined by users through \texttt{\_norm\_calc.py} in \texttt{utils}. The three types of spatial constraints handled by the algorithm are:

\begin{enumerate}
    \item \textbf{Region constrained: } This type of constraint arises when we can place either a \textit{maximum} of or \textit{exactly} $s$ sensors in a certain region, while the remaining $r-s$ sensors must be placed outside the constraint region.
    \begin{itemize}
        \item{\textbf{Maximum:}} This case deals with applications in which the number of sensors in the constraint region should be less than or equal to $s$. This functionality has been implemented through \texttt{max\_n} in \texttt{\_norm\_calc.py}. 
        \item{\textbf{Exact:}} This case deals with applications in which the number of sensors in the constraint region should equal $s$. This functionality has been implemented through \texttt{exact\_n} in \texttt{\_norm\_calc.py}. 
     \end{itemize}
    \item \textbf{Predetermined:} This type of constraint occurs when a certain number of sensor locations $s$ are already specified, and  optimized locations for the remaining sensors are desired. This functionality has been implemented through \texttt{predetermined} in \texttt{\_norm\_calc.py}. 
    \item \textbf{Distance constrained:} This constraint enforces a minimum distance $d$ between selected sensors. This functionality has been implemented through \texttt{distance} in \texttt{\_norm\_calc.py}. 
\end{enumerate}

\begin{lstlisting}[language=Python, basicstyle=\small\ttfamily,caption={Constrained sensor selection workflow: (1) Define circular constraint with center at (20,5) and radius 5, (2) Apply Generalized QR (GQR) optimization to select optimal sensor locations within the constraint, (3) Fit SSPOR model with SVD basis, and (4) Visualize selected sensors with constraint overlay.}, label={lst:circleSensors},captionpos=t]
circle = ps.utils._constraints.Circle(center_x = 20, center_y = 5, radius = 5, loc = 'in', data = X_train) 
circle.draw_constraint() 
circle.plot_constraint_on_data(plot_type='image')
circle.plot_grid(all_sensors=all_sensors)
const_idx, rank = circle.get_constraint_indices(all_sensors = all_sensors_unconst,info= X_train)
s = 4
optimizer = ps.optimizers.GQR()
optimizer_kwargs={'idx_constrained':const_idx,
         'n_sensors': r,
         'n_const_sensors': s,
         'all_sensors':all_sensors,
         'constraint_option':"exact_n"}
basis = ps.basis.SVD(n_basis_modes=n_sensors)
model = ps.SSPOR(basis = basis, optimizer = optimizer, n_sensors = r)
model.fit(X_train,**optimizer_kwargs)
top_sensors = model.get_selected_sensors()
dataframe = circle.sensors_dataframe(sensors = top_sensors)
circle.plot_constraint_on_data(plot_type='image')
circle.plot_selected_sensors(sensors = top_sensors, all_sensors = all_sensors)
circle.annotate_sensors(sensors = top_sensors, all_sensors = all_sensors)
\end{lstlisting}

\begin{figure}[t!]
	\centering
	\begin{subfigure}{.2\textwidth}
		\includegraphics[width=\textwidth]{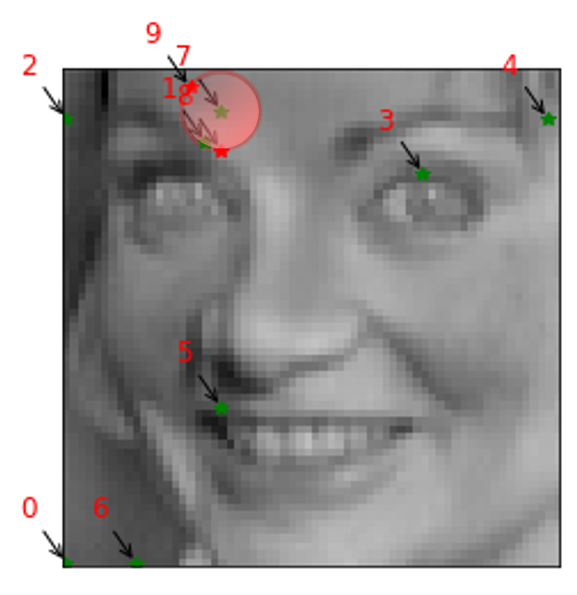}
		\caption{\label{fig:circle}Exact (s=4) }
	\end{subfigure}
	\begin{subfigure}{.21\textwidth}
		\includegraphics[width=\textwidth]{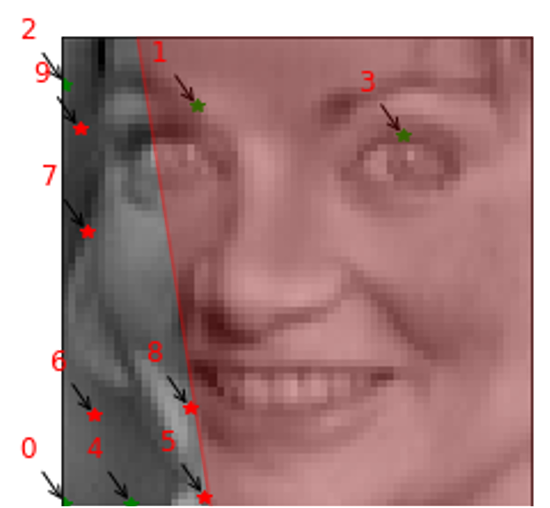}
		\caption{\label{fig:line}Exact (s = 2)}
	\end{subfigure}
	\begin{subfigure}{.2\textwidth}
		\includegraphics[width=\textwidth]{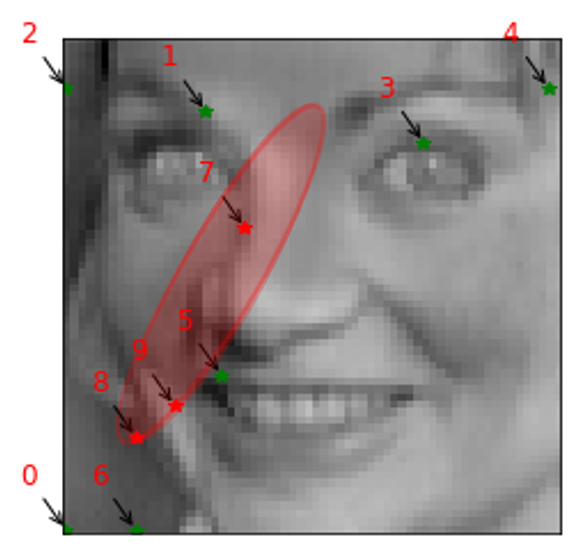}
		\caption{\label{fig:ellipse}Exact (s= 3)}
	\end{subfigure}
 \begin{subfigure}{.2\textwidth}
		\includegraphics[width=\textwidth]{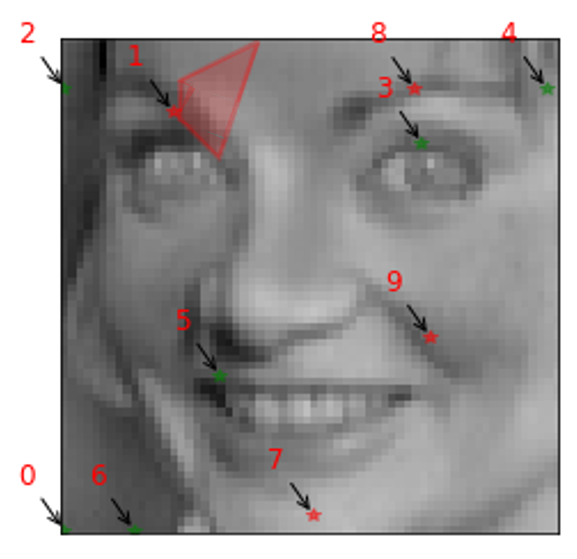}
		\caption{\label{fig:polygon}Exact (s=0)}
	\end{subfigure}
	\begin{subfigure}{.2\textwidth}
		\includegraphics[width=\textwidth]{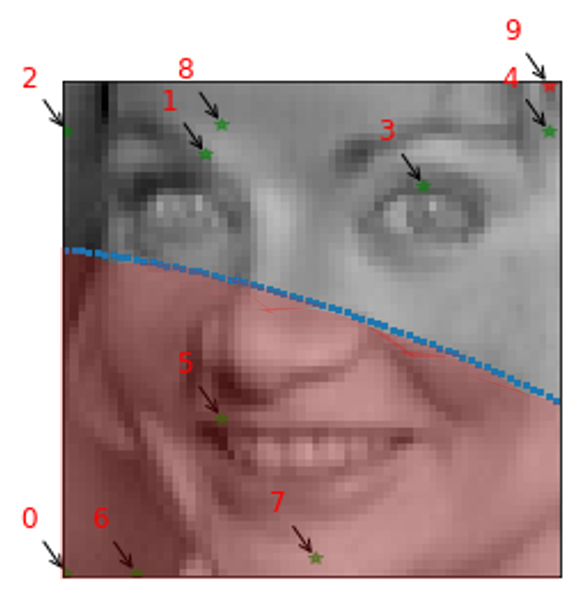}
		\caption{\label{fig:parabola}Exact (s= 4)}
	\end{subfigure}
\begin{subfigure}{.2\textwidth}
    \includegraphics[width=\textwidth]{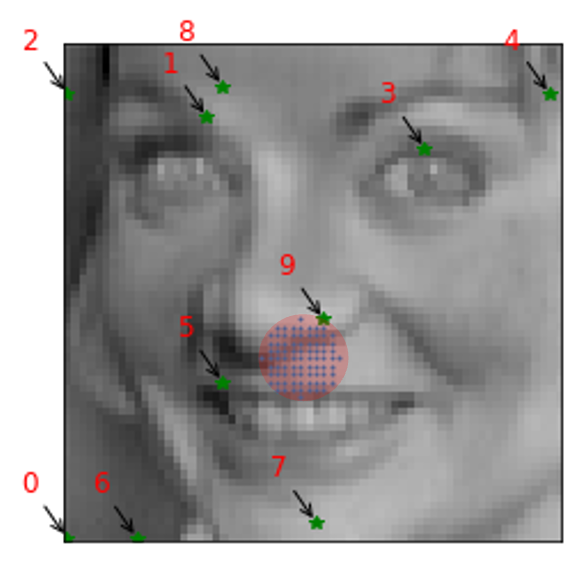}
    \caption{\label{fig:custom}Exact (s=0)}
\end{subfigure}
\begin{subfigure}{.41\textwidth}
    \includegraphics[width=\textwidth]{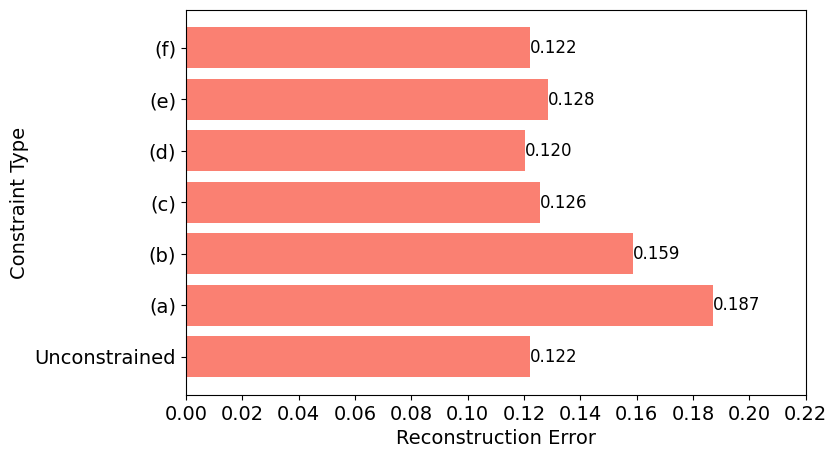}
    \caption{\label{fig:custom}Constraints increase error}
\end{subfigure}

	\caption{\label{fig:Olivetti}\textbf{Constraint region definition, visualization and plotting capabilities of the algorithm.} The algorithm incorporates various geometric constraint regions including circles (a), lines (b), ellipses (c),  polygons (d), parabolas (e), that adapt to user's specific requirements. Comprehensive visualization capabilities are integrated directly into the codebase, with green stars indicating original QR sensor positions and red stars highlighting sensors that have been repositioned to satisfy constraints. This visual distinction enables users to make more informed decisions. Additionally, the system supports user-defined custom functions (f) for creating specialized constraint regions tailored to specific applications. Reconstruction error on the test set increases when sensors are constrained to specific locations (g), with crowded sensor configurations exhibiting higher errors due to inter-sensor interference.}
\end{figure}

To implement spatial constraints in our sensor placement methodology, we must first define the shape of our constraint region and then identify which grid points fall within the designated constraint regions. We accomplish this using the specialized classes and functions provided in the \texttt{\_constraints.py} module in \texttt{utils}. The classes define constraint regions in various geometric shapes including \texttt{Circle}, \texttt{Ellipse}, \texttt{Polygon}, \texttt{Parabola}, \texttt{Line}, and \texttt{Cylinder}. Additionally, we provide \texttt{UserDefinedConstraints} options that allow users to supply their own constraints either as a Python \texttt{(.py)} file containing their constraint shape definition or as an equation string. These classes work with two distinct data formats: image data defined by pixel coordinates and tabular data stored in dataframes with explicit x, y, and z values. Upon instantiation of the class, users must provide the input \texttt{data} in one of two formats: a matrix representation for images or a structured dataframe for tabular data. In the latter case, the class requires additional keyword arguments \texttt{(kwargs)} specifying the columns corresponding to \texttt{X\_axis}, \texttt{Y\_axis}, \texttt{Z\_axis}, and \texttt{Field} parameters, thereby enabling the extraction of relevant data from their respective columns for subsequent analysis.

Once an instance of these classes have been initiated, as seen in the example code snippet above \autoref{lst:circleSensors}, the functions \texttt{draw\_constraint()}, \texttt{ plot\_constraint\_on\_data()} and \texttt{plot\_grid()} provide visualization of the constraint region, the constraint region on a snapshot of the data and all possible sensor locations respectively. The function \texttt{get\_constraint\_indices()} is used to identify sensors within constrained regions across multiple data formats and returns indices of sensors located within the specified spatial constraints. These can then be used in subsequent optimization routines to ensure that sensor placement adheres to physical or practical limitations.

The code snippet demonstrates how \texttt{General QR (GQR)} optimizer can then be used to handle spatial constraints with the \texttt{constraint\_option} = \texttt{exact\_n}, which ensures exactly $s = 4$ sensors are placed within the constrained region. The \texttt{optimizer\_kwargs} dictionary passes crucial parameters: constrained indices \texttt{(idx\_constrained)}, total number of sensors \texttt{(n\_sensors)}, number of constrained sensors \texttt{(n\_const\_sensors)}, total available sensor locations \texttt{(all\_sensors)} and the constraint option. The \texttt{SSPOR} model integrates the basis, optimizer and the model is fit on the input data \texttt{X\_train} with the specified optimizer key word arguments, yielding an optimal sensor placement that balances reconstruction accuracy with the imposed spatial constraints.

Following model optimization, the algorithm identifies the optimal sensor placement using the function \texttt{get\_selected\_sensors()}, which returns a subset of sensors \texttt{(top\_sensors)} determined to be most effective for the given constraints. These sensors can subsequently be extracted into a structured dataframe via the \texttt{sensors\_dataframe()} method for detailed analysis. The spatial distribution of constraints is then visualized through the \texttt{plot\_constraint\_on\_data()} method, which generates a representation of the constraint boundaries superimposed on the underlying data field. To facilitate comparative assessment, the \texttt{plot\_selected\_sensors()} method graphically differentiates between the optimal sensor locations and the complete sensor array, while the \texttt{annotate\_sensors()} method applies appropriate metadata labels to each sensor position, enabling efficient interpretation of the optimization results. Figure~\ref{fig:Olivetti} illustrates the diverse constraint geometries, constraint typologies, and visualization capabilities implemented in PySensors 2.0, demonstrating the framework's enhanced flexibility for sensor placement optimization. 

\subsection{Sensor Landscapes and the TPGR Optimizer}
The D-optimal objective in Eqn.~\ref{eqn:detobj} suffers from two limitations: it is not defined for the under-sampling case $p<r$, and it is hard to interpret and visualize directly. Ref.~\cite{klishin2023data} resolves these limitations by adding a prior regularization and decomposing the resulting objective into sums over the placed sensors:
\begin{equation}
    \mathcal{H}\equiv -\log \det(\mathbf{S}^{-2}+(\Sel \bTi_r)^T(\Sel \bTi_r)/\eta^2)\approx E_b+\sum\limits_{i\in \gamma}h_i+\sum\limits_{i\neq j \in\gamma}J_{ij},
    \label{eqn:twopoint}
\end{equation}
where $\mathbf{S}$ is the assumed prior covariance matrix of the coefficients $\ba$ and $\eta$ is the assumed sensor noise magnitude. The typical prior covariances are $\mathbf{S}\propto \mathbf{I}$ (isotropic Gaussian) or $\mathbf{S}=\mathbf{\Sigma_r}/\sqrt{N}$ (normalized singular values of training data). The series expansion over terms with more sensors is in principle exact, but we approximate it with the first two terms. In the approximation, $E_b$ is a constant term that does not affect sensor selection, $h_i,J_{ij}$ are the interaction landscapes computed from the basis and the prior. The objective in Eqn.~\ref{eqn:twopoint} involves only summation over the selected sensors and is thus cheaper to evaluate and update than the original objective in Eqn.~\ref{eqn:detobj}.

The approximate objective is used in the Two Point GReedy (TPGR) optimizer that can return a user-specified number of sensors $p$ for any mode number $r$. In contrast, the QR algorithm returns exactly $p=r$ sensors in order of decreasing importance through pivoting, and all following sensors are random. The sensor sets returned by TPGR are nearly equivalent to QR for isotropic prior $\mathbf{S}\propto \mathbf{I}$ and small noise $\eta$.

After a SSPOR model is fit using the TPGR Optimizer, the one point and two point energy landscapes can be computed as seen in~\autoref{lst:TPGRsnippet}. For prior values, the input can either be a numpy array, corresponding to the diagonal part of the covariance matrix (e.g. all equal for an isotropic prior), or the string 'decreasing', which computes the normalized singular values from the training data.
\begin{lstlisting}[language=Python, basicstyle=\small\ttfamily, caption={Implementation of the Two Point Greedy Optimizer}, label={lst:TPGRsnippet},captionpos=t]
    basis = ps.basis.SVD(n_basis_modes=r)
    optimizer = ps.optimizers.TPGR(n_sensors, noise, prior)
    model = ps.SSPOR(basis=basis, optimizer=optimizer)
    model.fit(data)
    sensors=model.get_selected_sensors()
    one_pt_landscape = model.one_pt_energy_landscape(prior, noise)
    two_pt_landscape = model.two_pt_energy_landscape(prior, noise, sensors)
\end{lstlisting}
\subsection{Reconstruction Solvers}
Once the set of $p$ sensors has been determined using any of the methods, the sensor measurements $\by$ can be used to determine the state coefficients $\ba$. Under the assumption of linearity, the reconstruction always takes the shape $\hat{\ba}=\bA\by$ for some matrix $\bA:r\times p$.

The first version of the reconstruction matrix corresponds to the Least Squares solution via the Moore-Penrose pseudoinverse:
\begin{equation}
    \bA_{LS}=(\Sel \bTi_r)^\dagger.
\end{equation}

PySensors 2.0 adds the Regularized Least Squares solution derived in Ref.~\cite{klishin2023data}:
\begin{equation}
    \bA_{RLS}=\left( \mathbf{S}^{-2}+(\Sel \bTi_r)^T(\Sel \bTi_r)/\eta^2 \right)^{-1} (\Sel \bTi_r)^T/\eta^2,
\end{equation}
where similarly to the TPGR optimizer, $\mathbf{S}$ is the assumed prior covariance matrix and $\eta$ is the assumed sensor noise magnitude. The relative magnitude of the coefficient prior and the noise determines whether the reconstruction primarily relies on the measurements or the prior information. This Regularized Least Squares solution is now the default reconstruction solver for PySensors 2.0.
\begin{lstlisting}[language=Python, basicstyle=\small\ttfamily, caption={Reconstruction using Regularized Least Squares}, label={reglsq}]
    model.predict(x_test, prior, noise=noise)
\end{lstlisting}
The 'unregularized' method will use the Least Squares method using Moore-Penrose pseudoinverse reconstruction solver.
\begin{lstlisting}[language=Python, basicstyle=\small\ttfamily]
    model.predict(x_test, method='unregularized')
\end{lstlisting}
\subsection{Uncertainty Quantification and Heatmaps}
For any choice of the reconstruction matrix $\bA$, the reconstruction $\hat{\ba}=\bA\by$ predicts the \emph{most likely} coefficients of the reconstructed state, from which the full state can be obtained via projection $\hat{\bx}=\bTi_r\hat{\ba}$. However, the reconstruction is sensitive to the sensor noise. The expected error of the reconstruction is captured by the covariance matrix derived in Ref.~\cite{klishin2023data}:
\begin{align}
    \mathbf{K}=\bTi_r \mathbf{B} \mathbf{B}^T \bTi_r^T;\quad \mathbf{B}=\eta  \bA. \label{eqn:K}
\end{align}


In practice, the whole covariance matrix in Eqn.~\ref{eqn:K} is both hard to store as it takes $n\times n$ space in computer memory, and hard to interpret as it has a low rank. Instead, PySensors 2.0 computes a reduced metric: the \emph{uncertainty heatmap} $\sigma=\sqrt{diag(\mathbf{K})}$ or the marginal standard deviation of each pixel of the reconstruction.
Note that for linear reconstructions, the error metrics do not depend on the system state, and thus can be computed with fairly fast numerical linear algebra in time that does not scale with the number of train or test snapshots $N$.

\begin{lstlisting}
    x_test = X_test[:, sensors]
    predicted_state = model.predict(x_test, prior, noise)
    sigma = model.std(prior, noise)
\end{lstlisting}

\section{Examples}
In this section, we demonstrate the enhanced capabilities of PySensors 2.0 through illustrative examples that highlight key advancements in the framework. We present case studies focused on three critical areas: constraint-aware optimization for realistic deployment scenarios using the \texttt{GQR} optimizer, one and two point sensor landscapes, and rigorous uncertainty quantification with the TPGR optimizer for sensor networks subject to measurement noise. For additional functionalities readers are directed to the examples available on the PySensors documentation site~\footnote[1]{https://github.com/dynamicslab/pysensors}.

\subsection{Functional constraints for a Nuclear simulation Example}
 Consider the challenge of monitoring a test capsule with an electric heater at the center that mimics the neutronics effect of a nuclear fuel rod. The goal is to study heat transfer from the rod to circulating water before reactor installation. Since the test capsule is axisymmetric, computational fluid dynamics (CFD) simulations model half the domain with the heater at the center. The goal is to strategically place point thermocouples within the capsule to capture the temperature profile. For additional details on this experiment, see Section 4.3 ``Steady-state simulation of the OPTI-TWIST prototype'' in Karnik et al. \cite{karnik2024constrained}.  
 \begin{lstlisting}[language=Python, basicstyle=\small\ttfamily, caption={Implementation of Sparse Sensor Placement Optimization for Reconstruction (SSPOR) using PySensors. The code demonstrates setting up a rank-5 SVD basis, QR optimizer, and fitting the model to identify optimal sensor locations from the available data. The model selects r=5 sensors from all possible sensor locations to minimize reconstruction error.},label={lst:QR},captionpos=t]
    r = 5
    basis = ps.basis.SVD(n_basis_modes=r)
    optimizer  = ps.optimizers.QR()
    model = ps.SSPOR(basis=basis, optimizer=optimizer, n_sensors=r)
    model.fit(data)
    all_sensors = model.get_all_sensors()
    sensors = model.get_selected_sensors()
 \end{lstlisting}
 \begin{figure}[t!]
 \centering
 \begin{subfigure}{0.2\textwidth}
		\includegraphics[width=\textwidth]{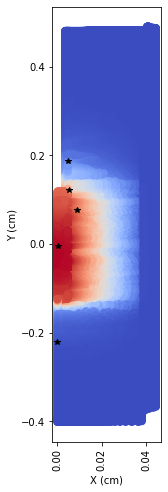}
    \caption{\label{fig:unconstrained}}
	\end{subfigure}
 \begin{subfigure}{0.2\textwidth}
		\includegraphics[width=\textwidth]{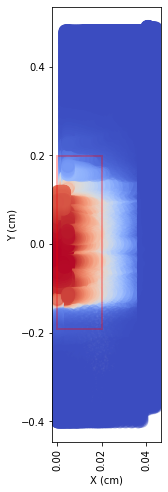}
		\caption{\label{fig:constrainedondata}}
	\end{subfigure}
	\begin{subfigure}{0.292\textwidth}
		\includegraphics[width=\textwidth]{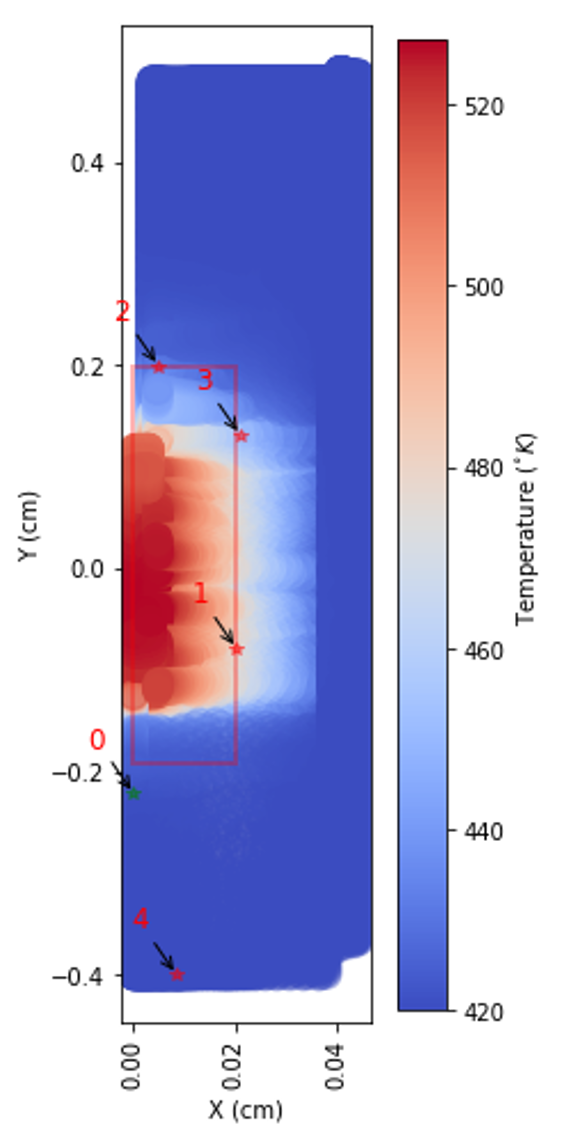}
		\caption{\label{fig:constrained}}
	\end{subfigure}
\caption{\label{fig:NuclearExample} (a) Unconstrained sensor placement optimization results in placing sensors in heater adjacent locations. Implementation of spatial constraints includes defining an exclusion region around the heater, plotting the  constraint region overlaid on experimental data (b), and final constrained sensor locations positioned away from the heater (c). }
\end{figure}
 The proximity to the heater element creates significant space constraints that render sensor placement in heater-adjacent locations experimentally infeasible. We first look at the placement of $r = 5$ unconstrained sensors using \texttt{QR} as the optimizer, and the \texttt{SVD} as the \texttt{basis} as seen in~\autoref{lst:QR}. \autoref{fig:unconstrained} shows that 3 of the 5 sensors are adjacent to the heater (black stars). Based on this we want to constrain the domain around the heater between $x_1 = 0, x_2 = 0.02, y_1 = -0.19, y_2 = 0.19$ cm to have no sensors.
 
A polygon class is initialized to identify sensor locations within the specified rectangular domain as shown in~\autoref{lst:GQR}. A polygonal constraint region is first defined using four corner coordinates $(x_1, y_1), (x_2, y_1), (x_2, y_2),(x_1, y_2)$, with the constraint configured to operate on interior points (\texttt{loc = in}). As the data is stored within a dataframe (df), this case differs from \autoref{sec:HardConstraints}, as we now provide \texttt{data = df} and must specify the \texttt{Y\_axis = 'Y (m)', X\_axis = 'X (m)'} and \texttt{Field = 'Temperature (K)'}, as opposed to working with images where spatial coordinates are implicit. The column names \texttt{'X (m)'}, \texttt{'Y (m)'}, and \texttt{'Temperature (K)'} correspond to the X-axis, Y-axis, and the temperature variable measured by the sensors, respectively.
 \begin{lstlisting}[language=Python, basicstyle=\small\ttfamily, caption={ Constrained sensor placement optimization using PySensors with a rectangular polygon constraint. The code defines spatial boundaries, applies the General QR (GQR) optimizer with SVD basis to select r optimal sensors within the constraint region, and visualizes the results with sensor annotations on the temperature field contour map.},label={lst:GQR},captionpos=t]
 polygon = ps.utils._constraints.Polygon([(x1, y1),(x2,y1),(x2,y2),(x1, y2)],loc = 'in', data = df, Y_axis = 'Y (m)', X_axis = 'X (m)', Field = 'Temperature (K)')
polygon.draw_constraint(plot = (fig,ax)) 
polygon.plot_constraint_on_data(plot_type='contour_map', plot = (fig,ax), s = 20) 
const_idx, rank = polygon.get_constraint_indices(all_sensors=all_sensors, info =df) 
s = 0
optimizer = ps.optimizers.GQR()
optimizer_kwargs={'idx_constrained':const_idx,
         'n_sensors':r,
         'n_const_sensors':s,
         'all_sensors':all_sensors,
         'constraint_option':"max_n"}
basis = ps.basis.SVD(n_basis_modes=r)
model = ps.SSPOR(basis = basis, optimizer = optimizer, n_sensors = r)
model.fit(data,**optimizer_kwargs)
top_sensors = model.get_selected_sensors() 
data_sens = polygon.sensors_dataframe(sensors = top_sensors) 
polygon.plot_constraint_on_data(plot_type='contour_map',plot = (fig,ax), s = 20) 
polygon.annotate_sensors(sensors = top_sensors, all_sensors=all_sensors) 
 \end{lstlisting}

The constraint visualization is implemented through two complementary approaches. First, the geometric boundary of the constraint region is rendered onto the specified figure and axis objects through \texttt{draw\_constraint}. Next, an overlay visualization of the constraint region onto the underlying data field using contour mapping is achieved by the function \texttt{plot\_constraint\_on\_data} as shown in \autoref{fig:constrainedondata}.
The constraint indices are extracted from the complete sensor array using the \texttt{get\_constraint\_indices} method, which returns both the constrained sensor indices and their corresponding rank ordering. The optimization framework utilizes the General QR (\texttt{GQR}) algorithm configured with constraint-aware parameters, including the constrained sensor indices \texttt{const\_idx}, total number of sensors \texttt{r} and number of constrained sensors \texttt{s}. As we want zero sensors in the constrained region, both \texttt{max\_n} and \texttt{exact\_n} will give us the same sensor configuration in this case.

The \texttt{Sparse Sensor Placement Optimization for Reconstruction (SSPOR)} model combines \texttt{SVD} basis functions with the \texttt{GQR} optimizer to identify near-optimal sensor locations within the constrained domain. After training on the dataset with specified optimizer keyword arguments (\texttt{kwargs}), the model extracts the highest-ranked sensor positions through \texttt{get\_selected\_sensors}. They can be formatted into a structured dataframe revealing the locations of the sensors using the function \texttt{sensors\_dataframe}. The visualization pipeline overlays these optimized sensor locations on the constraint region's contour map through\texttt{ plot\_constraint\_on\_data} and \texttt{annotate\_sensors} as seen in \autoref{fig:constrained}. The plot annotates sensors with red stars for locations that changed from the unconstrained case and green stars for locations that remained the same. 

Thus, based on spatial constraints and their geometric shapes that might arise in engineering applications, the framework can initialize various constraint classes and obtain the corresponding constraint indices. The system handles data in both dataframe and image formats, making it applicable to simulation and experimentation data.
Sensor location visualization reveals how the constrained optimization shifts sensor positions from their unconstrained near-optimal configuration. Additional analysis tools include reconstruction error studies and noise-induced uncertainty heatmaps that quantify the impact of these spatial constraints on sensing performance \cite{karnik2024constrained} as described in the next section.

\subsection{Two-point Greedy Optimizer Example} We use the SST dataset to showcase the \texttt{TPGR} optimizer, which uses two-point greedy optimization introduced in Ref.~\cite{klishin2023data} for sensor placement.
\begin{figure}[t!]
    \begin{subfigure}[t]{0.5\textwidth}
		\includegraphics[width=\textwidth]{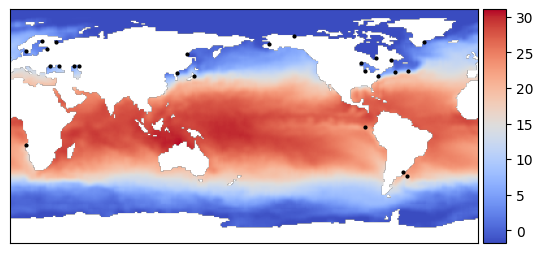}
		\caption{Learned Sensor Locations\label{fig:TPGRsensors}}
	\end{subfigure}
    \begin{subfigure}[t]{0.5\textwidth}
		\includegraphics[width=\textwidth]{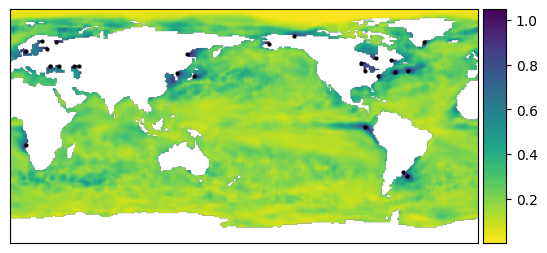}
		\caption{Uncertainty Heatmap \label{fig:UQHeatmap}}
	\end{subfigure}
    \begin{subfigure}[t]{0.5\textwidth}
		\includegraphics[width=\textwidth]{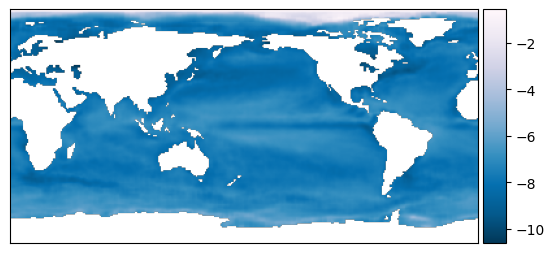}
		\caption{One Point Energy Landscape \label{fig:OnePtLandscape}}
	\end{subfigure}
    \begin{subfigure}[t]{0.5\textwidth}
		\includegraphics[width=\textwidth]{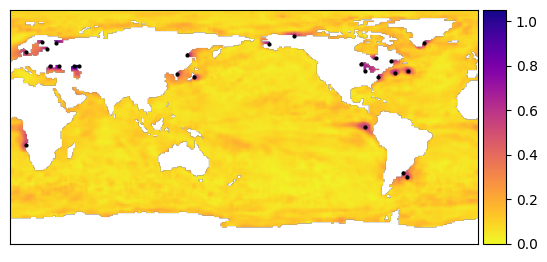}
		\caption{Two Point Energy Landscape \label{fig:TwoPtLandscape}}
	\end{subfigure}
    \caption{Sensor placement and reconstruction using Two-Point Greedy Optimization results in (a) learned sensor locations on a snapshot of the SST dataset (units in degrees Celcius), (b) Uncertainty heatmap of the reconstructed state, (c) One-point energy landscape at each location, (d) Two-point energy landscape at each location, as a sum of two point interactions with all placed sensors. \label{fig:TPGRExample}}
\end{figure}
\begin{lstlisting}[language=Python, basicstyle=\small\ttfamily]
    r = 100
    prior = np.full(r, 1000)
    noise = 1
    p = 25
    model = ps.SSPOR(
        basis=ps.basis.SVD(n_basis_modes=r),
        optimizer=ps.optimizers.TPGR(n_sensors=p, prior, noise)
    )
    model.fit(data_train)
\end{lstlisting}
We place $p=25$ sensors using \texttt{TPGR} as the \texttt{optimizer} and \texttt{SVD} as the \texttt{basis}, as shown in \autoref{fig:TPGRsensors}. For the \texttt{TPGR} optimizer to place sensors, prior and noise are required arguments. In this example, we are using a Isotropic Gaussian prior (flat prior). Other options of prior can be used: such as a prior constructed from normalized singular values of the training set of the data (decreasing  prior), truncated to the number of basis modes $r$, or a user defined prior.

Next, we compute the uncertainty heatmap for the reconstructed state as shown in \autoref{fig:UQHeatmap}. Using the \texttt{TPGR} optimizer enables us to calculate the one point energy landscape as shown in \autoref{fig:OnePtLandscape} and two point energy landscapes. Both energy landscape computations require prior and noise as an input. Two point landscape also requires the selected sensors whose two-point interactions need to be computed. If the selected sensors are an array of multiple sensors, the two point energy landscape will be the sum of two point interactions of all the selected sensors as shown in \autoref{fig:TwoPtLandscape}. If the selected sensors is just a single sensor, the two point energy landscape will be the two point interaction of that particular sensor.
\begin{lstlisting}[language=Python, basicstyle=\small\ttfamily]
    sensors = model.get_selected_sensors()
    data_test = data_test[:, sensors]
    predicted_state = model.predict(data_test, prior, noise)
    sigma = model.std(prior)
    one_pt_landscape = model.one_pt_energy_landscape(prior, noise)
    two_pt_landscape = model.two_pt_energy_landscape(prior, noise, selected_sensors)
\end{lstlisting}
\subsection{Reconstruction Comparison Example}
\begin{figure}[t!]
\captionsetup[subfigure]{justification=Centering}
    \begin{subfigure}[t]{0.5\textwidth}
		\includegraphics[width=\textwidth]{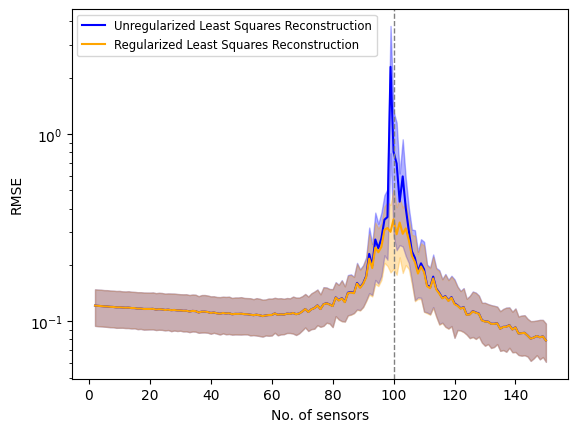}
		\caption{Flat prior sensor placement,\\ flat prior reconstruction \label{fig:flatflatOlivetti}}
	\end{subfigure}
    \begin{subfigure}[t]{0.5\textwidth}
		\includegraphics[width=\textwidth]{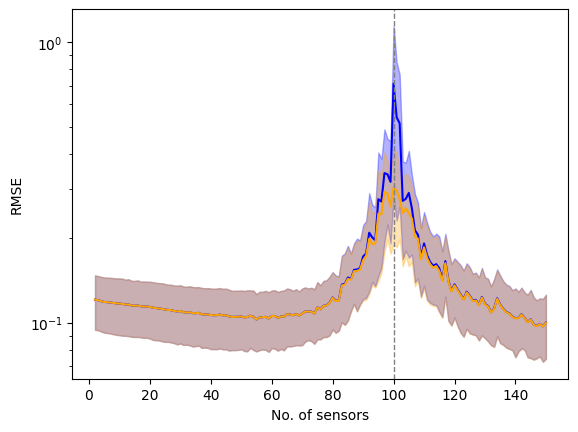}
		\caption{Decreasing prior sensor placement, \\ flat prior reconstruction \label{fig:decreasingflatOlivetti}}
	\end{subfigure}
    \begin{subfigure}[t]{0.5\textwidth}
		\includegraphics[width=\textwidth]{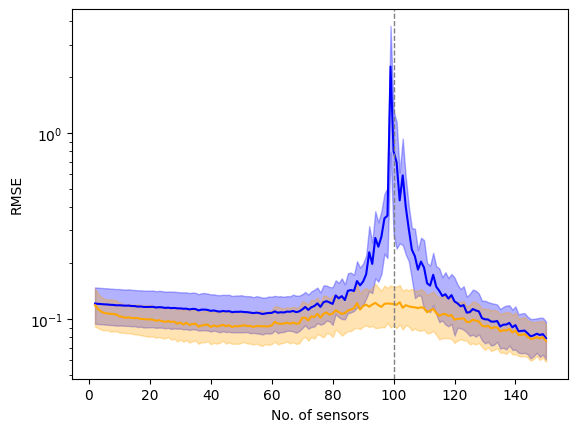}
		\caption{Flat prior sensor placement, \\ decreasing prior reconstruction \label{fig:flatdecreasingOlivetti}}
	\end{subfigure}
    \begin{subfigure}[t]{0.5\textwidth}
		\includegraphics[width=\textwidth]{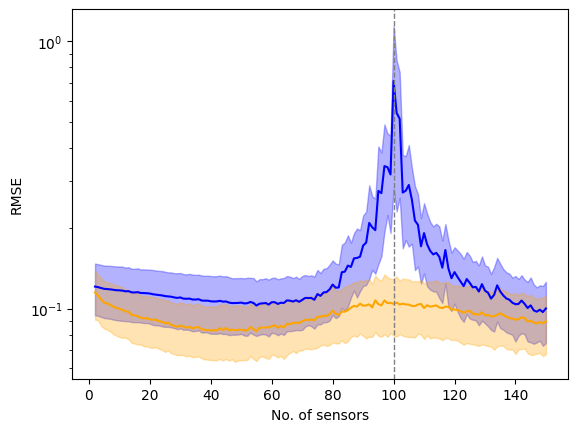}
		\caption{Decreasing prior sensor placement, \\ decreasing prior reconstruction \label{fig:decreasingdecreasingOlivetti}}
	\end{subfigure}
    \caption{Reconstruction error curves for different choices of the sensor placement prior and reconstruction prior on the Olivetti dataset. \label{fig:ReconstructionComparison}}
\end{figure}
In another notebook, we compare the reconstruction methods on the Olivetti dataset: unregularized reconstruction and regularized reconstruction. We use the \texttt{TPGR} optimizer for sensor placement using either a flat prior or a decreasing prior. We then reconstruct using both unregularized and regularized reconstruction, and compare their RMSE values with the true states along a range of number of placed sensors, from 1 to 150 sensors.  Since regularized reconstruction also requires a prior, there are four possible combinations that need to be considered: \autoref{fig:flatflatOlivetti} shows sensor placement using a flat prior \texttt{TPGR} optimizer and a flat prior for the regularized reconstruction. In \autoref{fig:decreasingflatOlivetti}, sensors are placed using the decreasing prior and regularized reconstruction is done using a flat prior. \autoref{fig:flatdecreasingOlivetti} uses flat prior for sensor placement and decreasing prior for regularized reconstruction. Finally, \autoref{fig:decreasingdecreasingOlivetti} uses a decreasing prior for both sensor placement and regularized reconstruction. The flat prior selected for this particular dataset is the scaled identity $4I$.

\begin{figure}[t!]
    \centering
    \includegraphics[width=0.7\linewidth]{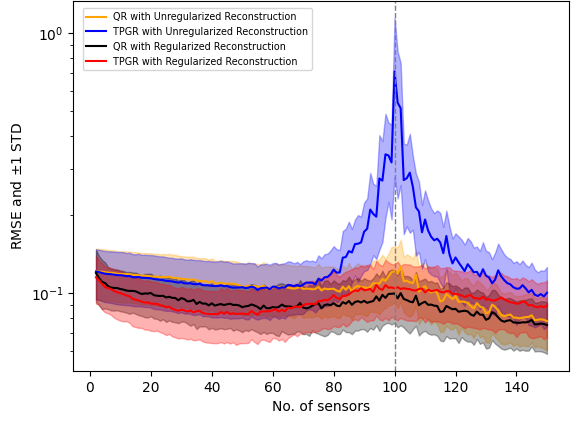}
    \caption{Comparison of sensor placement using QR and TPGR optimizers for unregularized and regularized reconstruction on the Olivetti Dataset\label{fig:TPGRQR}}
\end{figure}

We then compare the reconstruction methods for the \texttt{QR} and \texttt{TPGR} optimizers, using only the decreasing prior for sensor placement and regularized reconstruction. \autoref{fig:TPGRQR} shows a comparison of four plots, \texttt{QR} with unregularized reconstruction, \texttt{QR} with decreasing prior regularized reconstruction, decreasing prior \texttt{TPGR} with unregularized reconstruction and decreasing prior \texttt{TPGR} with decreasing prior regularized reconstruction. Empirically, using a regularized reconstruction with a decreasing prior reduces the RMSE significantly.
\section{Practical Tips}
In this section, we delineate key practical considerations for optimizing sensor placements using \texttt{PySensors 2.0}. Our discussion focuses primarily on the reconstruction methodology, as this domain encompasses the most significant enhancements implemented in the latest version. 

The selection of an appropriate basis is a critical determinant in both classification and reconstruction processes, particularly when dealing with high-dimensional systems. For static data or image analysis, the Singular Value Decomposition (SVD) basis often proves optimal as it efficiently captures the maximum variance within the dataset. Conversely, for time-dependent phenomena, Dynamic Mode Decomposition (DMD) or Fourier-based approaches may yield superior results by inherently accounting for temporal evolution patterns.
The enhanced custom basis functionality in \texttt{PySensors 2.0} now facilitates seamless integration with specialized basis types, such as DMD modes derivable from complementary packages like \texttt{PyDMD}~\cite{ichinaga2024pydmd}, as demonstrated in the basis example notebook available in the \texttt{PySensors} repository.
\begin{figure}[t!]
    \centering
    \includegraphics[width=\linewidth]{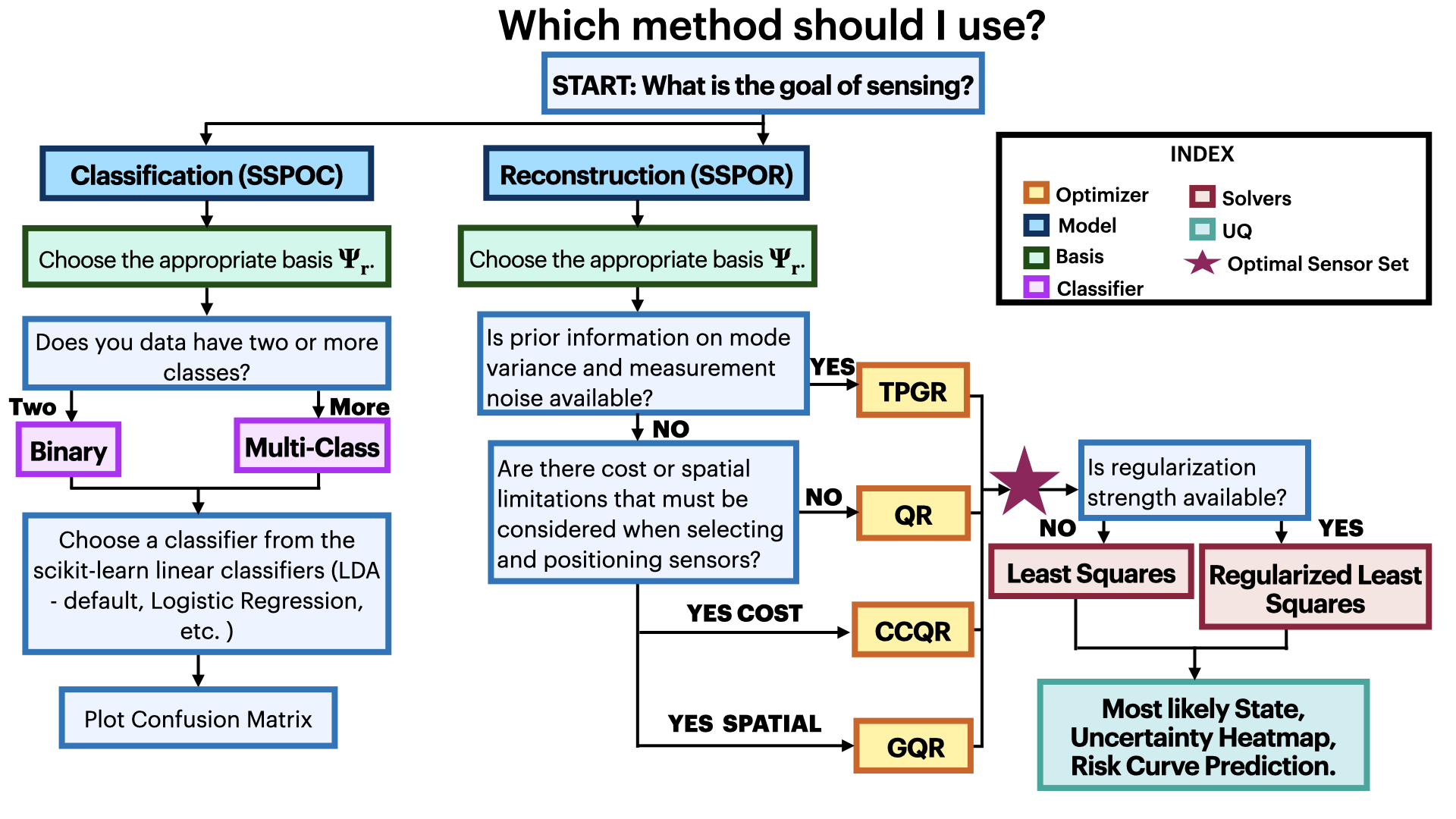}
    \caption{When selecting a sensing method in PySensors, consider your primary objective:
For field reconstruction in standard settings, use QR with Identity or SVD basis. For classification tasks, leverage SVD basis with SSPOC optimizer. When facing spatial constraints, choose GQR optimizer. For under-sampling $(p < r)$ and over-sampling cases $(p > r)$ scenarios , select TPGR optimizer. In noisy environments enable uncertainty quantification for robust results.}
    \label{fig:practical_use}
\end{figure}

Generally, reconstruction error is expected to diminish as both sensor quantity $p$ and basis mode count $r$ increase. However, the relative proportion between sensors and modes represents a crucial design parameter. When employing an SVD basis, two counterintuitive phenomena emerge.  First, as mode count increases beyond a certain threshold, additive measurement noise begins to dominate the error profile, paradoxically increasing overall reconstruction error (see~\cite{peherstorfer2020stability}). In the previous version of \texttt{Pysensors}, this effect could be mitigated by oversampling i.e. deploying more sensors than modes which would lead to randomized sensor selection once sensors exceeded modes $p > r$~\cite{clark2020multi}. Second, the reconstruction tends to be particularly unstable when the number of sensors matches the number of modes $p\approx r$ since in that regime the reconstruction requires inverting a particularly ill-conditioned matrix. This phenomenon is known as \emph{double descent} across machine learning literature. At fixed mode number $r$, the reconstruction error first decreases with number of sensors $p$, then spikes at $p\approx r$, then decreases again. In practical terms, two techniques reduce the reconstruction error: using a regularized reconstruction (\autoref{fig:practical_use}) and deliberately \emph{under-}sampling ($p<r$) the number of sensors for reconstruction. Before deploying a set of sensors optimized with \texttt{PySensors}, we recommend constructing the RMSE curves and deciding on the values of $p,r$ that lead to the best compromise between reconstruction error and deployment cost (not necessarily $p=r$). A detailed theory of double descent in sensors, including its origins and mitigation, is subject of a forthcoming publication \cite{klishin2025doubledescent}.

When implementing spatial constraints, it is optimal to maintain $p = r$ for sensor placement and then evaluate the best number of sensors for reconstruction. Exceeding this threshold causes \texttt{GQR} to place sensors randomly, resulting in loss of control over sensor placement within the constrained region. This random allocation risks potential selection of sensors within the constrained region when $p > r$. 

In conclusion, \autoref{fig:practical_use} presents a comprehensive flowchart for near-optimal sensor placement using \texttt{PySensors 2.0}. Users can select a basis for their data, and then determine whether to place sensors for reconstruction or classification purposes.
For reconstruction applications, different optimizers are available based on specific requirements such as cost limitations, spatial constraints, or prior knowledge of the data. The current default method for reconstruction is regularized least squares, where users can either provide their own prior or allow \texttt{PySensors} to calculate it automatically. Alternatively, users can opt for the standard least squares solver if regularization is not desired.
After obtaining the optimal sensor configuration, we recommend analyzing the placement effectiveness through uncertainty heatmaps and reconstruction error metrics to validate the solution for your specific application.

\section{Future Functionality}
Future work to enhance \texttt{PySensors} functionality includes integrating the two-point greedy (TPGR) method for sensor placement with additional cost landscapes and spatial constraints. This will prove beneficial in both under- and over-sampling scenarios where spatial constraints limit sensor placement options, a common challenge in most engineering applications. Another critical development involves incorporating all uncertainty sources beyond measurement noise into the UQ heatmap and risk curve predictions. This comprehensive uncertainty quantification would help explain the double descent phenomenon observed in sensor-based reconstruction, where reconstruction error initially decreases with additional sensors but then increases due to overfitting or noise amplification.
\section{Acknowledgments}
The authors acknowledge support from the Boeing Company, NSF AI Institute in Dynamic Systems under grant 2112085 and through the Idaho National Laboratory (INL) Laboratory Directed Research \& Development (LDRD) Program under DOE Idaho Operations Office Contract DE-AC07-05ID14517 for LDRD-22A1059-091FP.
\printbibliography
\end{document}